\documentclass[10pt,twocolumn,letterpaper]{article}

\usepackage[pagenumbers]{cvpr} 

\usepackage{graphicx}
\usepackage{amsmath}
\usepackage{amssymb}
\usepackage{booktabs}
\usepackage{bbding}
\usepackage{multirow}
\usepackage{float}
\usepackage{xcolor}
\usepackage{colortbl,booktabs}
\usepackage{color}
\usepackage[pagebackref,breaklinks,colorlinks]{hyperref}

\usepackage[capitalize]{cleveref}
\crefname{section}{Sec.}{Secs.}
\Crefname{section}{Section}{Sections}
\Crefname{table}{Table}{Tables}
\crefname{table}{Tab.}{Tabs.}

\definecolor{tablegreen}{RGB}{113,198,113}
\definecolor{tablered}{RGB}{205,51,51}

\definecolor{linkcolor}{HTML}{ED1C24}
\definecolor{baselinecolor}{gray}{.9}
\newcommand{\baseline}[1]{\cellcolor{baselinecolor}{#1}}

\def\cvprPaperID{5763} 
\def\confName{CVPR}
\def\confYear{2022}

\begin{document}

\title{Real-time Object Detection for Streaming Perception}

\author{Jinrong Yang$^{1}$,~~~ Songtao Liu$^2$\thanks{Corresponding author},~~~ Zeming Li$^2$,~~~ Xiaoping Li$^1$,~~~ Jian Sun$^2$\\
$^1$Huazhong University of Science and Technology, $^2$Megvii Technology\\
\tt\small yangjinrong@hust.edu.cn;liusongtao@megvii.com;lizeming@megvii.com;
\\
\tt\small lixiaoping@hust.edu.cn;sunjian@megvii.com}

\maketitle
\begin{abstract}
   Autonomous driving requires the model to perceive the environment and (re)act within a low latency for safety. While past works ignore the inevitable changes in the environment after processing, streaming perception is proposed to jointly evaluate the latency and accuracy into a single metric for video online perception. In this paper, instead of searching trade-offs between accuracy and speed like previous works, we point out that endowing real-time models with the ability to predict the future is the key to dealing with this problem. We build a simple and effective framework for streaming perception. It equips a novel Dual-Flow Perception module (DFP), which includes dynamic and static flows to capture the moving trend and basic detection feature for streaming prediction. Further, we introduce a Trend-Aware Loss (TAL) combined with a trend factor to generate adaptive weights for objects with different moving speeds. Our simple method achieves competitive performance on Argoverse-HD dataset and improves the AP by 4.9\% compared to the strong baseline, validating its effectiveness. Our code will be made available at \url{https://github.com/yancie-yjr/StreamYOLO}.
   
\end{abstract}

\section{Introduction}
\label{sec:intro}

\begin{figure}[t]      
\captionsetup[subfigure]{skip = 0pt}      
\centering      
\subfloat[baseline]{\includegraphics[width =1.5in]{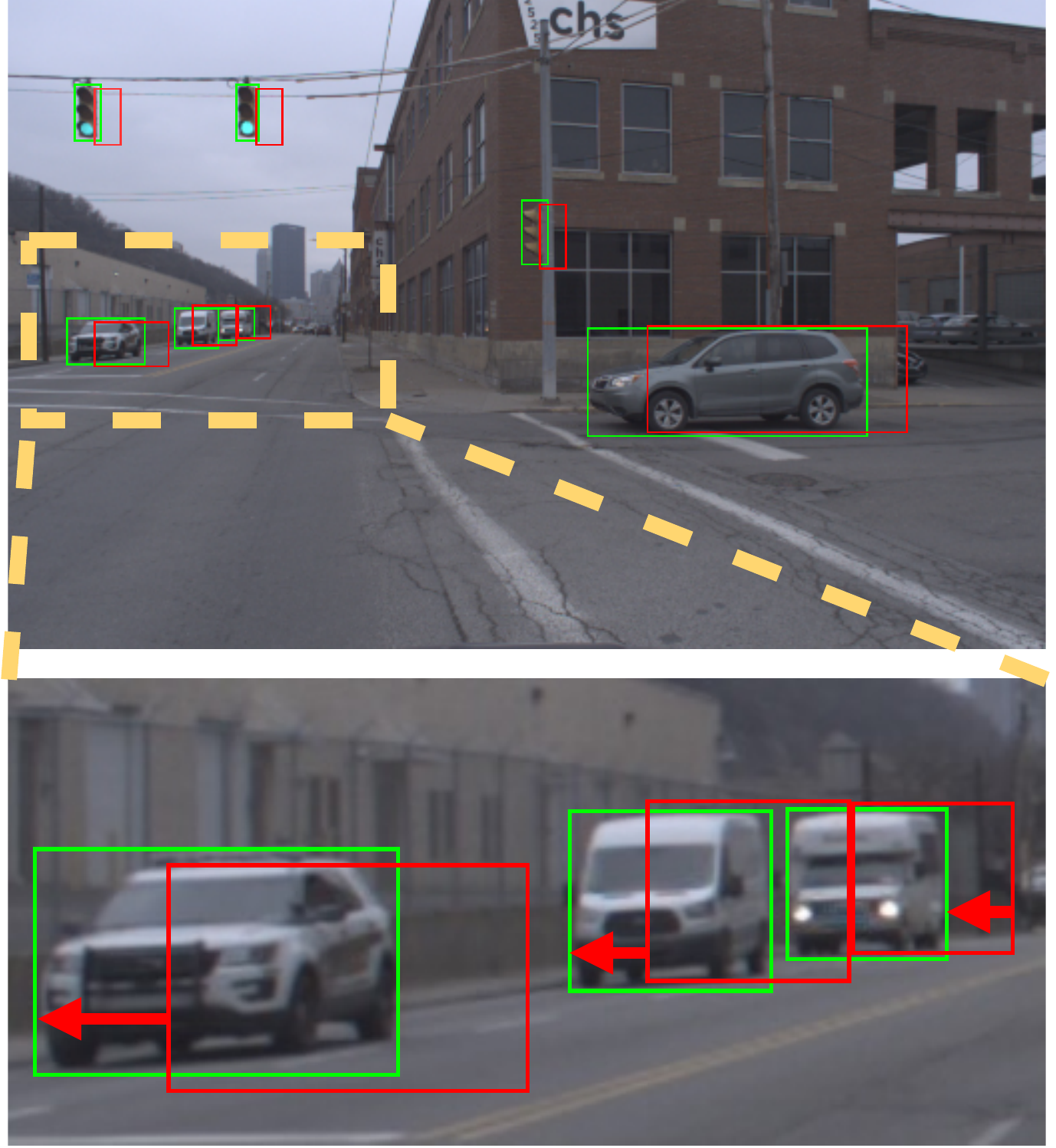}
\label{fig:baseline}}\ \      
\subfloat[ours]
{\includegraphics[width = 1.5in]{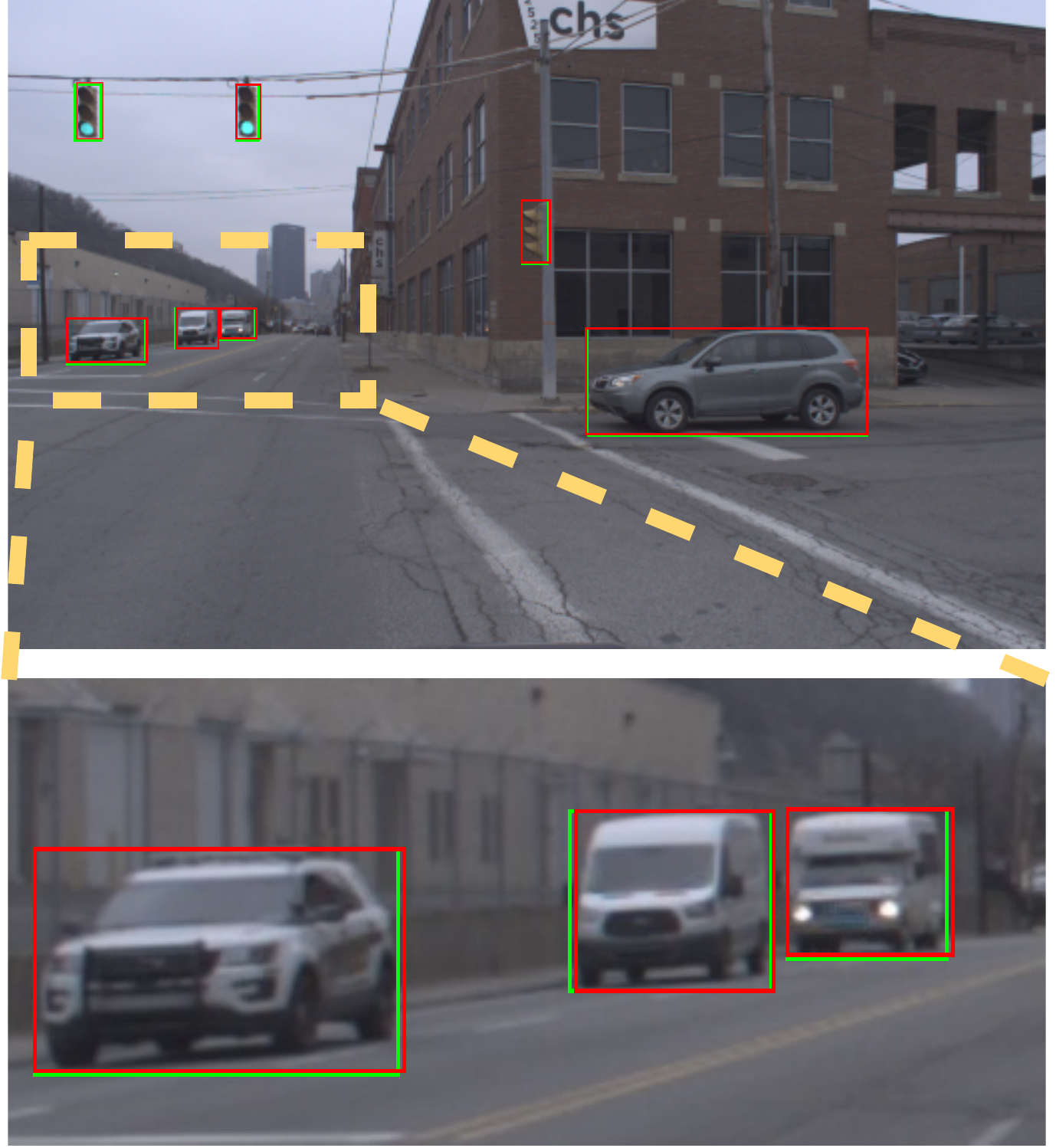}
\label{fig:ours}}\ \      
\caption{Illustration of visualization results of base detector and our method. The green boxes are ground truth, while the red ones are predictions. The red arrows mark the shifts of the prediction boxes caused by the processing time delay while our approach alleviates this issue.}      
\label{fig:fig1}      
\vspace{-0.1in}  
\end{figure}

One critical factor for autonomous safe driving is to perceive its environment and (re)act within a low latency. 
Recently, several real-time detectors \cite{yolo1,yolo2,yolo3,yolo4,yolo5,yolox,rfb,asff} achieve competitive performance under the low latency restriction. But they are still explored in an \emph{offline} setting \cite{streamer}. In a real-world vision-for-online scenario, no matter how fast the model becomes, the surrounding environment has changed once the model finishes processing the latest frame. As shown in Fig.~\ref{fig:fig1}(a), the inconsistency between perceptive results and the changed state may trigger unsafe decisions for autonomous driving. Thus for online perception, detectors are imposed to have the ability of future forecasting. 

To tackle this issue, \cite{streamer} firstly proposes a new metric named streaming accuracy, which integrates latency and accuracy into a single metric for real-time online perception. It jointly evaluates the output of the entire perception stack at every time instant, forcing the perception to forecast the state where the model finishes processing. 
With this metric, \cite{streamer} shows a significant performance drop of several strong detectors \cite{htc,retinanet,maskrcnn} from offline setting to streaming perception. Further, \cite{streamer} proposes a meta-detector named Streamer that can incorporate any detector with decision-theoretic scheduling, asynchronous tracking, and future forecasting to recover much of the performance drop. 
Following this work, Adaptive streamer\cite{adaptivestreamer} adopts numerous approximate executions based on deep reinforcement learning to learn a better trade-off online. 
These works focus on searching for a better trade-off policy between speed and accuracy for some existing detectors, while a novel streaming perception model design is not well studied. 

One more thing ignored by the above works is the existing real-time object detectors \cite{yolo5, yolox}. By strong data augmentation and delicate architecture design, they achieve competitive performance and can run faster than 30 FPS.
With these ''fast enough'' detectors, there is no space for accuracy and latency trade-off on streaming perception as the current frame results from the detector are always matched and evaluated by the next frame. These real-time detectors can narrow the performance gap between streaming perception and offline settings. In fact, both the 1st \cite{yoloxx} and 2nd \cite{2nd} place solution of Streaming Perception Challenge (Workshop on Autonomous Driving at CVPR 2021) adopt real-time models YOLOX \cite{yolox} and YOLOv5 \cite{yolo5} as their base detectors. 
Standing on the shoulder of the real-time models, we find that now the performance gap all comes from the fixed inconsistency between the current processing frame and the next matched frame. Thus the key solution for streaming perception is to predict the results of the \emph{next} frame at the \emph{current} state. 

Unlike the heuristic methods such as Kalman filter \cite{kalman} adopted in \cite{streamer}, in this paper, we directly endow the real-time detector with the ability to predict the future of the next frame. Specifically, we construct triplets of the last, current, and next frame for training, where the model gets the last and current frames as input and learns to predict the detection results of the next frame. We propose two crucial designs to improve the training efficiency: i) For model architecture, we conduct a Dual-Flow Perception (DFP) module to fuse the feature map from the last and current frames. 
It consists of a dynamic flow and a static flow. Dynamic flow pays attention to the moving trend of objects for forecasting while static flow provides basic information and features of detection through a residual connection. 
ii) For the training strategy, we introduce a Trend Aware Loss (TAL) to dynamically assign different weights for localizing and forecasting each object, as we find that objects within one frame may have different moving speeds. 

We conduct comprehensive experiments on Argoverse-HD\cite{argoverse,streamer} dataset, showing significant improvements in the stream perception task. 
In summary, the contributions of this work are as three-fold as follows:
\begin{itemize}
\item With the strong performance of real-time detectors, we find the key solution for streaming perception is to predict the results of the \emph{next} frame. This simplified task is easy to be structured and learned by a model-based algorithm.  

\item We build a simple and effective streaming detector that learns to forecast the next frame. We propose two adaptation modules, \emph{i.e.}, Dual-Flow Perception (DFP) and Trend Aware Loss (TAL), to perceive the moving trend and predict the future.      

\item We achieve competitive performance on Argoverse-HD\cite{argoverse,streamer} dataset without bells and whistles. Our method improves the mAP by +4.9\% compared to the strong baseline of the real-time detector and shows robust forecasting under the different moving speeds of the driving vehicle. 
\end{itemize}

\section{Related Works}
\label{sec:related}

\paragraph{Image object detection.} In the era of deep learning, detection algorithms can be split into the two-stage \cite{fastrcnn,fpn,fasterrcnn,liu2019pay,wu2020multi,zheng2020cross} and the one-stage \cite{ssd,retinanet,fcos,yolo1,rfb,adaptivenms,iqdet,autoassign,borderdet,lla} frameworks. Some works, such as YOLO series \cite{yolo1,yolo2,yolo3,yolo4,yolo5,yolox}, adopt a bunch of training and accelerating tricks to achieve strong performance with real-time inference speed. Our work is based on the recent real-time detector YOLOX \cite{yolox} which achieves strong performance among real-time detectors. 

\paragraph{Video object detection.} Streaming perception also relates to video object detection. Some recent methods\cite{mega,rdn,fgfa,tracker} employ attention mechanism, optical flow, and tracking method,  aiming to aggregate rich features for the complex video variation, \emph{e.g.}, motion blur, occlusion, and out-of-focus. However, they all focus on the offline setting, while streaming perception considers the online processing latency and needs to predict the future results.

\paragraph{Video prediction.} Video prediction tasks aim to predict the results for the unobserved future data. Current tasks include future semantic/instance segmentation. For semantic segmentation, early works \cite{predicting,bayesian} construct a mapping from past segmentation to future segmentation. Recent works \cite{segmenting,vsaric2019single,warp,predictive} convert to predict intermediate segmentation features by employing deformable convolutions, teacher-student learning,  flow-based forecasting, LSTM-based approaches, etc. For instance segmentation prediction, some approaches predict the pyramid features\cite{luc2018predicting} or the feature of varied pyramid
levels jointly\cite{sun2019predicting,apanet}. The above prediction methods do not consider the misalignment of prediction and environment change caused by processing latency, leaving a gap to real-world application. In this paper, we focus on the more practical task of streaming perception. 


\paragraph{Streaming perception.} Streaming perception task coherently considers latency and accuracy. \cite{streamer} firstly proposes sAP to evaluate accuracy under the consideration of time delay. Facing latency, non-real-time detectors will miss some frames. \cite{streamer} proposes a meta-detector to alleviate this problem by employing Kalman filter\cite{kalman}, decision-theoretic scheduling, and asynchronous tracking\cite{tracker}. \cite{adaptivestreamer} lists several factors (e.g., input scales, switchability of detectors, and scene aggregation.) and designs a reinforcement learning-based agent to learn a better combination for a better trade-off. Fovea\cite{fovea} employs a KDE-based mapping to raise the upper limit of the offline performance. In this work, instead of searching better trade-off or enhancing base detector, we simplify the steaming perception to the task of ``predicting the next frame'' by a real-time detector. 



\begin{figure}[t]
\begin{center}
\includegraphics[width=\linewidth]{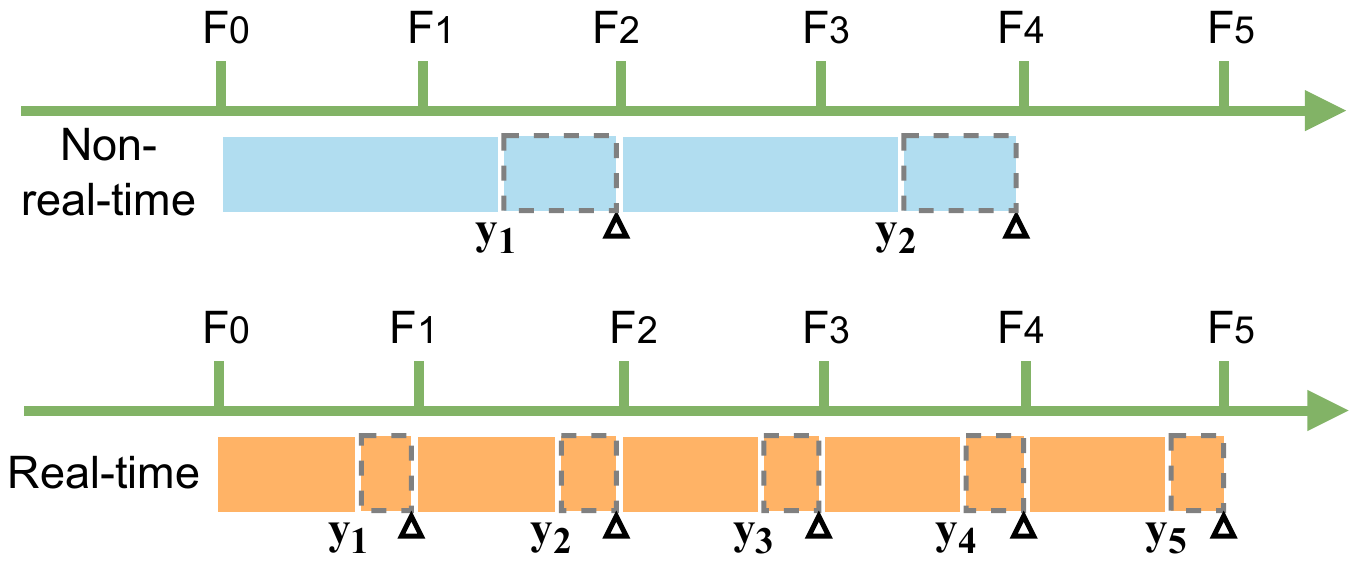}
\caption{Comparison on different detectors in streaming perception evaluation framework. Each block represents the process of the detector for one frame and its length indicates the running time. The dashed block indicates the time until the next frame data is received.}
\label{fig:streaming}
\end{center}
\end{figure}

\begin{figure}[b]
\begin{center}
\includegraphics[width=\linewidth]{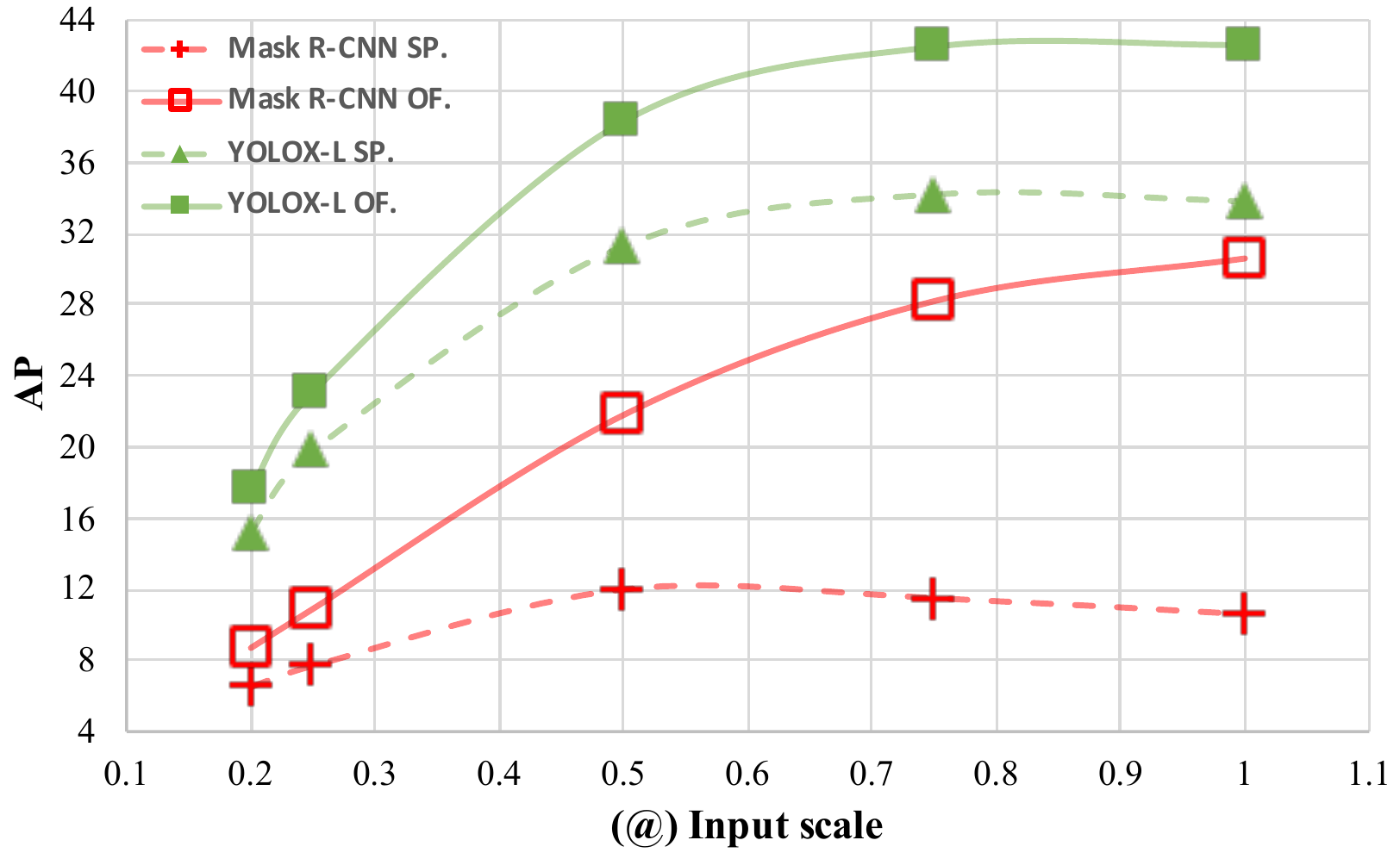}
\caption{The performance gap between \emph{offline} and streaming perception setting brings about on Argoverse-HD dataset. 'OF' and 'SP' indicate \emph{offline} and streaming perception setting respectively. The number after @ is the input scale (the full resolution is $1200 \times 1920$).}
\label{fig:contrast}
\end{center}
\end{figure}

\begin{figure*}
\setlength{\abovecaptionskip}{-3pt}
\begin{center}
\includegraphics[width=\linewidth]{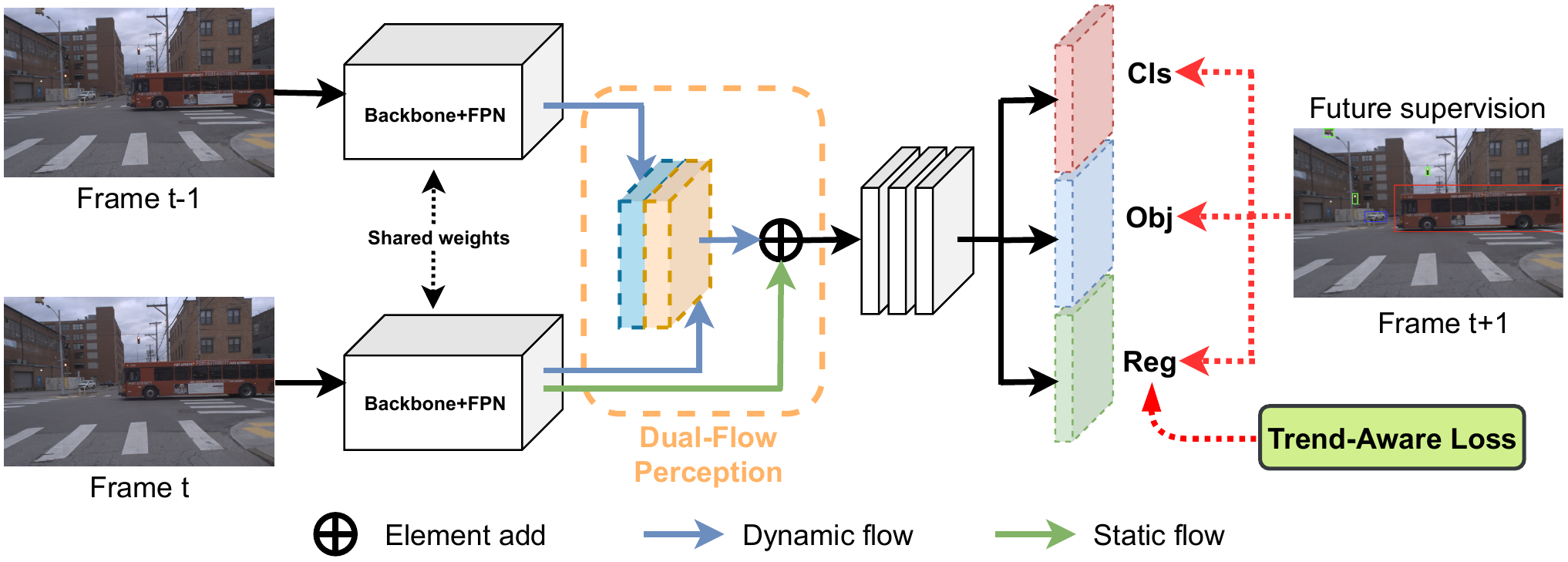}
\end{center}
\caption{The training pipeline. First, we adopt a shared weight CSPDarknet-53 with PANet to extract FPN features of the current and last image frames. Second, we use the proposed Dual-Flow Perception module (DFP) to aggregate feature maps and feed them to classification, objectness and regression head. Third, we directly utilize the ground truth of the next frame to conduct supervision. We also design a Trend-Aware Loss (TAL) applied to the regression head for efficient training.}
\label{fig:train}
\end{figure*}

\begin{figure}[b]
\begin{center}
\includegraphics[width=\linewidth]{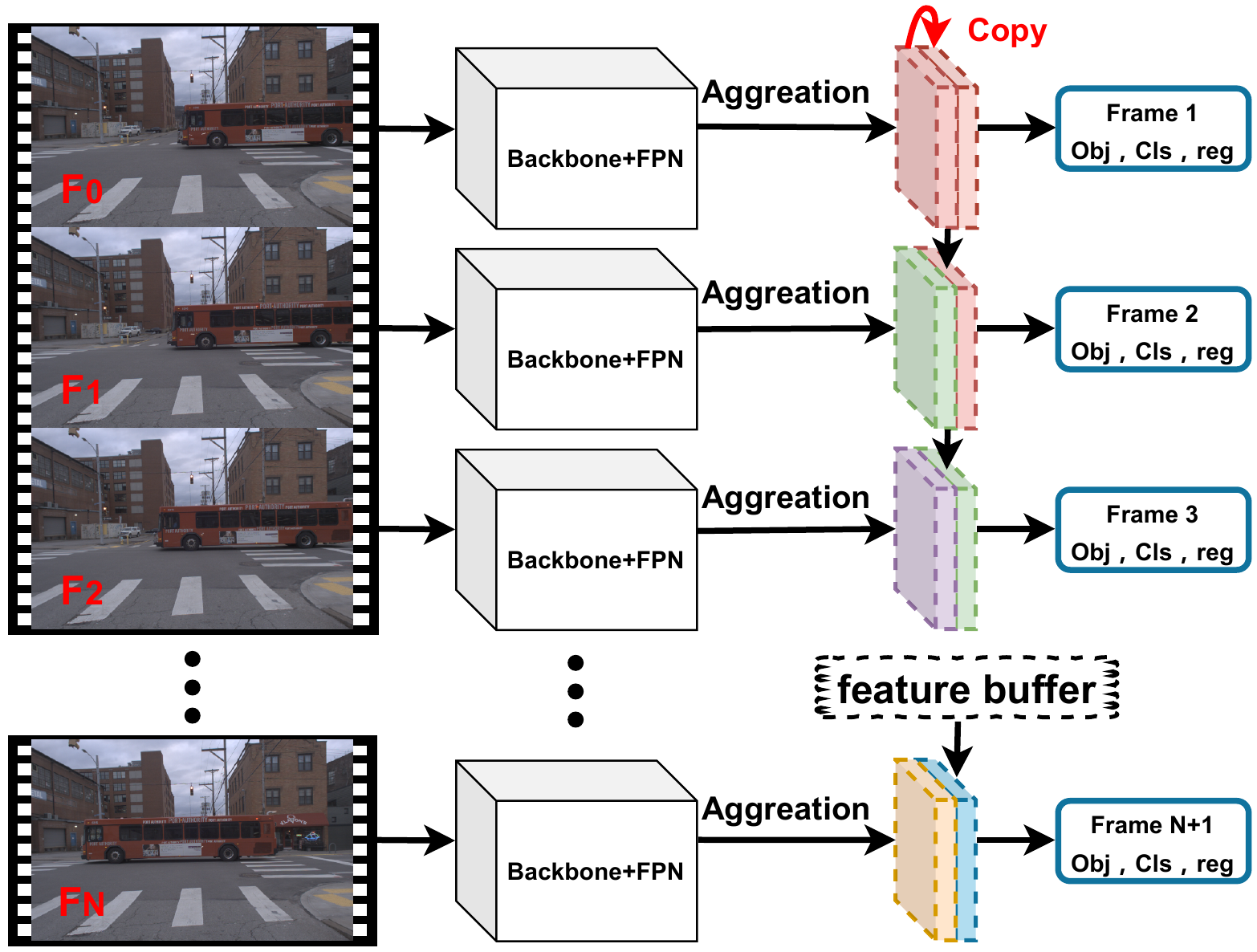}
\caption{The inference pipeline. We employ a feature buffer to save the historical features of the latest frame and thus only need to extract current features. By directly aggregating the features stored at the last moment, we save the time of handling the last frame again. For the beginning of the vedio, we copy the current FPN features as pseudo historical buffers to predict results.}
\label{fig:inference}
\end{center}
\end{figure}

\section{Methods}

\subsection{Streaming Perception}
Streaming perception organizes data as a set of sensor observations. To take the model processing latency into account, \cite{streamer} proposes a new metric named streaming AP (sAP) to simultaneously evaluate time latency and detection accuracy. As shown in Fig.~\ref{fig:streaming}, the streaming benchmark evaluates the detection results over a continuous time frame. After receiving and processing an image frame, sAP simulates the time latency among the streaming flow and examines the processed output with a ground truth of the actual world state.

For the example of a non-real-time detector, the output $y_1$ of the frame $F_1$ is matched and evaluated with the ground truth of $F_3$ and the result of $F_2$ is missed. Thus for the task of streaming perception, non-real-time detectors may miss many image frames and produce long-time shifted results, significantly hurting the performance of \emph{offline} detection.   

For real-time detectors (the total processing time of one image frame is less than the time interval of image streaming), the task of streaming perception becomes easy and clear. 
As we can see in Fig.~\ref{fig:streaming}, a real-time detector avoids the shifting problem with a fixed pattern of matching the next frame to the current prediction.
This fixed matching pattern not only eradicates the missed frames but also reduces the time shift for each matched ground truth. 

In Fig.~\ref{fig:contrast}, we compare two detectors, Mask R-CNN~\cite{maskrcnn} and YOLOX~\cite{yolox}, with several image scales and study the performance gap between streaming perception and offline settings. In the case of low-resolution input, the performance gaps are small for two detectors as they are all running in a real-time manner. However, with the resolution increasing, the performance drop of Mask R-CNN gets larger as it runs slower. For YOLOX, its inference speed maintains real-time with the resolution increasing, so that the gap is not correspondingly widened.     

\subsection{Pipeline}
\label{sec:3.2}

The fixed matching pattern from real-time detectors also enables us to train a learnable model to dig the potential moving trend and predict the objects of the next image frames. Our approach includes a basic real-time detector, an offline training schedule, an online inference strategy, which are described next.    

\paragraph{Base detector.} We choose the recent proposed YOLOX \cite{yolox} as our base detector. It inherits and carries forward YOLO series  \cite{yolo1,yolo2,yolo3} to an anchor-free framework with several tricks, \emph{e.g.}, decoupled heads \cite{head1,head2}, strong data augmentations \cite{mixup,copypaste}, and advanced label assigning \cite{ota},  achieving strong performance among real-time detectors.
It is also the 1st place solution \cite{yoloxx} of Streaming Perception Challenge in the Workshop on Autonomous Driving at CVPR 2021. Different from \cite{yoloxx}, we remove some engineering speedup tricks such as TensorRT and change the input scale to the half resolution ($600 \times 960$) to ensure the real-time speed without TensorRT. We also discard the extra datasets used in \cite{yoloxx}, \emph{i.e.}, BDD100K~\cite{yu2020bdd100k}, Cityscapes~\cite{cordts2016cityscapes}, and nuScenes~\cite{caesar2020nuscenes} for pre-training. These shrinking changes definitely decrease the detection performance compared to \cite{yoloxx}, but they alleviate the executive burden and allow extensive experiments. We believe the shrinking changes are orthogonal to our work and can be equipped to further improve the performance.       
\paragraph{Training.} We visualize our total training pipeline in Fig.~\ref{fig:train}. We construct the last, the current frames and next gt boxes to a triplet $(F_{t-1}, F_t, G_{t+1})$ for training. The main reason for this design is simple and direct: in order to predict the future position of objects, it is inevitable to know the moving status for each object. We thus take two adjacent frames ($F_{t-1}$, $F_{t}$) as input and train the model to directly predict the detection results of the next frame, supervised by the ground truth of $F_{t+1}$. Based on the triplets of inputs and supervision, we rebuild the training dataset to the formulate of $\{(F_{t-1}$, $F_{t}$, $G_{t+1})\}_{t=1}^{n_{t}}$, where $n_{t}$ is the total sample number. The first and last frame of each video streaming is excluded. With this rebuilt dataset, we can keep a random shuffling strategy for training and improve the efficiency with distributed GPU training as normal.


To better capture the moving trend between two input frames, we propose a Dual-Flow Perception Module (DFP) and a Trend-Aware Loss (TAL), introduced in the next subsection, to fuse the FPN feature maps of two frames and adaptively catch the moving trend for each object.

We also study another indirect task which parallelly predicts the current gt boxes $G_t$ and the offsets of object transformations from $G_t$ to $G_{t+1}$. However, according to some ablation experiments, described in the next section (Sec.~\ref{exp:pred}), we find that predicting the additional offsets always falls into a suboptimal task. One reason is that the value of the transformative offsets between two adjacent frames is small, involving some noise of numerical instability. It also has some bad cases where the label of the corresponding object is sometimes not reachable (new objects come or current objects disappear in the next frame).   

\paragraph{Inference.} The proposed model takes two image frames as input, bringing nearly twice computational cost and time consumption compared to the original detector. As shown in Fig.~\ref{fig:inference}, to eliminate the dilemma, we employ a feature buffer to store all the FPN feature maps of the previous frame $F_{t-1}$. At inference, our model only extracts the feature of the current image frame and then aggregates the historical features from the buffer. With this strategy, our model runs almost at the same speed as the base detector. For the beginning frame $F_0$ of the stream, we duplicate the FPN feature maps as pseudo historical buffers to predict results. This duplication actually means ``no moving'' status and the static results are inconsistent with $F_1$. Fortunately, the influence on performance is trivial as this case is rare.   

\subsection{Dual-Flow Perception Module (DFP)}
Given the FPN feature maps of the current frame $F_{t}$ and the historical frame $F_{t-1}$, we suppose two critical pieces of information the feature should have for predicting the next frame. One is the moving tendency to capture the moving state and estimate the magnitude of movement. The other is the basic semantic information for the detector to localize and classify the corresponding objects.

We thus design a Dual-Flow Perception (DFP) module to encode the expected features with the dynamic flow and static flow, as seen in Fig.~\ref{fig:train}. Dynamic flow fuses the FPN feature of two adjacent frames to learn the moving information. It first employs a shared weight $1\times1$ convolution layer followed by the batchnorm and SiLU \cite{swish} to reduce the channel to half numbers for both two FPN features. Then, it simply concatenates these two reduced features to generate the dynamic features. We have studied several other fusing operations like add, non-local block\cite{nonlocal}, STN\cite{stn} based on squeeze-and-excitation network\cite{senet}, where concatenation shows the best efficiency and performance (see Tab.~\ref{tab:operator} ). As for static flow, we reasonably add the original feature of the current frame through a residual connection. In the later experiments, we find the static flow not only provides the basic information for detection but also improves the predicting robustness across different moving speeds of the driving vehicle. 

\subsection{Trend-Aware Loss (TAL)}
We notice an important fact in streaming perception, in which the moving speed of each object within one frame is quite different. The variant trends come from many aspects: different sizes and moving states of their own, occlusions, or the different topological distances. 

Motivated by the observations, we introduce a Trend-Aware Loss (TAL) which adopts adaptive weight for each object according to its moving trend. Generally, we pay more attention to the fast-moving objects as they are more difficult to predict the future states. To quantitatively measure the moving speed, we introduce a trend factor for each object. We calculate an IoU (Intersection over Union) matrix between the ground truth boxes of $F_{t+1}$ and $F_{t}$ and then conduct the max operation on the dimension of $F_{t}$ to get the matching IoU of the corresponding objects between two frames. The small value of this matching IoU means the fast-moving speed of the object and vice versa. 
If a new object comes in $F_{t+1}$, there is no box to match it and its matching IoU is much smaller than usual. We set a threshold $\tau$ to handle this situation and formulate the final trend factor $\omega_i$ for each object in $F_{t+1}$ as:

\begin{equation}
    \label{eq:eq1}
    mIoU_i = \max_j (\{IoU(box^{t+1}_i,box^{t}_{j})\})
\end{equation}

\begin{equation}
    \label{eq:eq2}
    \omega_i = \left\{ {\begin{array}{*{20}{c}}
       {1/mIoU_i  \qquad mIoU_i \ge \tau }
    \\
       {1/\nu  \qquad \qquad mIoU_i < \tau }
\end{array}}, \right.
\end{equation}
where $\max_j$ represents the max operation among boxes in $F_t$, $\nu$ is a constant weight for the new coming objects. We set $\nu$ as 1.4 (bigger than 1) to reduce the attention according to hyper-parameters grid searching.

Note that simply applying the weight to the loss of each object will
change the magnitude of the total losses. This may disturb the balance between the loss of positive and negative samples and decrease the detection performance. Inspired by \cite{ioubb,ggiou}, we normalize $\omega_{i}$ to $\hat{\omega}_{i}$ intending to keep the sum of total loss unchanged: 

\begin{equation}
    \label{eq:eq3}
    \hat{\omega}_{i} = \omega _i\cdot\frac{{\sum_{i = 1}^N {{\mathcal{L}^{reg}_i}} }}{{\sum_{i = 1}^N {{\omega_i}}{\mathcal{L}^{reg}_i}}},
\end{equation}
where $\mathcal{L}^{reg}_i$ indicates the regression loss of object $i$. Next, we re-weight the regression loss of each object with $\hat{\omega}_{i}$ and the total loss is exhibited as:  

\begin{equation}
    \label{eq:eq4}
    {\mathcal{L}_{total}} = \sum\limits_{i \in positive}\hat{\omega}_{i}\mathcal{L}^{reg}_i  + {\mathcal{L}_{cls}} + {\mathcal{L}_{obj}}.
\end{equation}

\begin{table*}[t]
\vspace{-.2em}
\centering
\subfloat[
\textbf{Prediction task}. Comparisons on two types of prediction tasks.
\label{tab:Supervision}
]{
\scalebox{0.68}{
\begin{tabular}{@{}c|c|ccccc@{}}
\toprule
Method & sAP & $\rm sAP_{50}$ & $\rm sAP_{75}$\\
\midrule
Baseline & 31.2  & 54.8 & 29.5\\
Offsets & 31.0 \textcolor{tablered}{\small (\textbf{-0.2})} & 52.2 & 30.7\\
Next  & 34.2 \textcolor{tablegreen}{\small (\textbf{+3.0})} & 54.6 & 34.9\\
\bottomrule
\multicolumn{4}{c}{~}\\
\multicolumn{4}{c}{~}\\
\end{tabular}
} 
}
\centering
\hspace{1em}
\subfloat[
\textbf{Fusion feature}. Comparisons on three different patterns of features to fuse.
\label{tab:Features}
]{
\scalebox{0.68}{
\begin{tabular}{@{}c|c|cc|c@{}}
\toprule
Method & sAP & $\rm sAP_{50}$ & $\rm sAP_{75}$ & Latency\\
\midrule
Baseline & 31.2  & 54.8 & 29.5 & 18.23 ms\\
Input & 30.3 \textcolor{tablered}{\small (\textbf{-0.9)}} & 50.5 & 29.2 & 18.33 ms\\
Backbone & 30.5 \textcolor{tablered}{\small (\textbf{-0.7)}} & 50.5 & 30.5 & 18.76 ms\\
FPN & 34.2 \textcolor{tablegreen}{\small (\textbf{+3.0)}} & 54.6 & 34.9 & 18.98 ms\\
\bottomrule
\multicolumn{4}{c}{~}\\
\end{tabular}
} 
}
\centering
\hspace{1em}
\subfloat[
\textbf{Fusion operator}. Comparisons on different fusion operations.
\label{tab:operator}
]{
\scalebox{0.68}{
\begin{tabular}{@{}c|c|cc|c@{}}
\toprule
Operation & sAP & $\rm sAP_{50}$ & $\rm sAP_{75}$ & Latency\\
\midrule
Baseline & 31.2  & 54.8 & 29.5 & 18.23 ms\\
Add & 30.8 \textcolor{tablered}{\small (\textbf{-0.4)}} & 54.8 & 29.6 & 18.81 ms\\
NL & 32.7 \textcolor{tablegreen}{\small (\textbf{+1.5)}} & 56.1 & 30.7 & 26.11 ms\\
STN & 34.0 \textcolor{tablegreen}{\small (\textbf{+2.8)}} & 55.8 & 32.9 & 24.32 ms\\
Concatenation & 34.2 \textcolor{tablegreen}{\small (\textbf{+3.0)}} & 54.6 & 34.9 & 18.98 ms\\
\bottomrule
\end{tabular}
} 
}
\\
\centering

\caption{Ablation experiments for building a strong pipeline. We employ a basic YOLOX-L detector as the baseline for all experiments.}
\label{tab:ablations}
\end{table*}

\section{Experiments}
\label{sec:exp}


\subsection{Settings}

\paragraph{Datasets.} We conduct the experiments on video autonomous driving dataset Argoverse-HD \cite{argoverse,streamer} (High-frame-rate Detection),  which contains diverse urban outdoor scenes from two US cities. It has multiple sensors and high frame-rate sensor data (30 FPS).
Following \cite{streamer}, we only use the center RGB camera and the detection annotations provided by \cite{streamer}. We also follow the train/val split in \cite{streamer}, where the validation set contains 24 videos with a total of 15k frames.

\paragraph{Evaluation metrics.} We use sAP \cite{streamer} (the streaming perception challenge toolkit~\cite{sAP}) to evaluate all experiments. sAP is a metric for streaming perception. It simultaneously considers latency and accuracy. Similar to MS COCO metric \cite{coco}, it evaluates average mAP over IoU (Intersection-over-Union) thresholds from 0.5 to 0.95 as well as $\rm AP_{s}$, $\rm AP_{m}$, $\rm AP_{l}$ for small, medium and large object.

\paragraph{Implementation details.} If not specified, we use YOLOX-L \cite{yolox} as our default detector. All of our experiments are fine-tuned from the COCO pre-trained model by 15 epochs. We set batch size at 32 on 8 GTX 2080ti GPUs. We use stochastic gradient descent (SGD) for training. We adopt a learning rate of $0.001 \times BatchSize / 64$ (linear scaling\cite{linear}) and the cosine schedule with a warm-up strategy for 1 epoch. The weight decay is 0.0005 and the SGD momentum is 0.9. The base input size of the image is $600 \times 960$ while the long side evenly ranges from 800 to 1120 with 16 strides. We do not use any data augmentation (such as Mosaic\cite{yolo5}, Mixup\cite{mixup}, horizontal flip, etc.) since the feeding adjacent frames need to be aligned. For inference, we keep the input size at $600 \times 960$ and measure the processing time on a Tesla V100 GPU.

\subsection{Ablations for Pipeline}
We conduct ablation studies for the pipeline design on three crucial components: the task of prediction, the feature used for fusion, and the operation of fusion. We employ a basic YOLOX-L detector as the baseline for all experiments and keep the other two components unchanged when ablating one. In particular, all entries work in real-time (30 FPS) so that the comparison is fair.
\paragraph{Prediction task.}
\label{exp:pred}
We compare the two types of prediction tasks mentioned in Sec.~\ref{sec:3.2}. 
As shown in Tab.~\ref{tab:Supervision}, indirectly predicting current bounding boxes with corresponding offsets gets even worse performance than the baseline. In contrast, directly forecasting future results achieves significant improvement (+3.0 AP). This demonstrates the supremacy of directly predicting the results of the next frame.  

\paragraph{Fusion feature.} Fusing the previous and current information is important for the streaming task. For a general detector, we can choose three different patterns of features to fuse: input, backbone, and FPN pattern respectively. Technically, the input pattern directly concatenates two adjacent frames together and adjusts the input channel of the first layer. The backbone and FPN pattern adopt a $1\times1$ convolution followed by batch normalization and SiLU to reduce half channels for each frame and then concatenate them together. As shown in Tab.~\ref{tab:Features}. The results of the input and backbone pattern decrease the performance by 0.9 and 0.7 AP. By contrast, the FPN pattern significantly boosts 3.0 AP, turning into the best choice. These results indicate that the fusing FPN feature may get a better trade-off between capturing the motion and detecting the objects.

\paragraph{Fusion operation.} We also explore the fusion operation for FPN features. We seek several regular operators (\emph{i.e.}, element-wise add and concatenation) and advanced ones (\emph{i.e.}, spatial transformer network \cite{stn} (STN)\footnote{To implement STN for variable inputs, we adopt a SE \cite{senet} block to calculate the transformation parameters instead of using flatten operation and fully connected layers in the original STN.} and non-local network \cite{nonlocal} (NL)\footnote{For NL, we use the current feature to calculate the values and queries and use the previous feature to generate keys.}. Tab.~\ref{tab:operator} shows the performance among these operations. We can see that the element-wise add operation drops performance by 0.4 AP while other ones achieve similar gains. We suppose that adding element-wise values may break down the relative information between two frames and fail to learn trending information. And among effective operations, concatenation is prominent because of its light parameters and high inference speed.

\subsection{Ablations for DFP and TAL}

\paragraph{Effect of DFP and TAL.} To validate the effect of DFP and TAL, we conduct extensive experiments on YOLOX detectors with different model sizes. In Tab.~\ref{tab:table3}, ``Pipe.'' denotes our basic pipeline containing basic feature fusion and future prediction training. Compared to the baseline detector, the proposed pipelines have already improved the performance by 1.3 to 3.0 AP across different models. Based on these high-performance baselines, DFP and TAL can boost the accuracy of sAP by $\sim$1.0 AP independently, and their combinations further improve the performance by nearly 2.0 AP. These facts not only demonstrate the effectiveness of DFP and TAL but also indicate that the contributions of the two modules are almost orthogonal. 

Indeed, DFP adopts dynamic flow and static flow to extract the moving state feature and basic detection feature separately and enhances the FPN feature for streaming perception. Meanwhile, TAL employs adaptive weight for each object to predict different trending. We believe the two modules cover different points for streaming perception: architecture and optimization. We hope that our simple design of the two modules will lead to future endeavors in these two under-explored aspects.

\begin{table}[t]
\centering
\scalebox{0.75}{
\begin{tabular}{@{}c|ccc|c|c|cc@{}}
\toprule
Model & Pipe. & DFP & TAL & Off AP & sAP & $\rm sAP_{50}$ & $\rm sAP_{75}$\\
\midrule
\multirow{5}*{YOLOX-S} & & & & \multirow{5}*{32.0} & 26.3 & 48.1 & 24.0\\
~ & \checkmark & & & ~ & \baseline{$\rm 27.6_{\ \uparrow \ 1.3}$}  & \baseline{48.3} & \baseline{26.1}\\
~ &\checkmark &\checkmark & & ~ & 28.2 \textcolor{tablegreen}{\small (\textbf{+0.6)}} & 49.4 & 27.4\\
~ &\checkmark & &\checkmark & ~ & 28.1 \textcolor{tablegreen}{\small (\textbf{+0.5)}} & 49.1 & 27.0\\
~ &\checkmark &\checkmark &\checkmark & ~ & 28.8 \textcolor{tablegreen}{\small (\textbf{+1.2)}} & 50.3 & 27.6\\
\midrule
\multirow{5}*{YOLOX-M} & & & &\multirow{5}*{34.5} & 29.2 & 51.9 & 27.7\\
~ & \checkmark & & & ~ & \baseline{$\rm 31.2_{\ \uparrow \ 2.0}$} & \baseline{51.1} & \baseline{31.9}\\
~ &\checkmark &\checkmark & & ~ & 32.3 \textcolor{tablegreen}{\small (\textbf{+1.1)}} & 52.9 & 32.5\\
~ &\checkmark & &\checkmark & ~ & 31.8 \textcolor{tablegreen}{\small (\textbf{+0.6)}} & 53.1 & 31.8\\
~ &\checkmark &\checkmark &\checkmark & ~ & 32.9 \textcolor{tablegreen}{\small (\textbf{+1.7)}} & 54.0 & 32.5\\
\midrule
\multirow{5}*{YOLOX-L} & & & &\multirow{5}*{38.3} & 31.2 & 54.8 & 29.5\\
~ & \checkmark & & & ~ & \baseline{$\rm 34.2_{\ \uparrow \ 3.0}$} & \baseline{54.6} & \baseline{34.9}\\
~ &\checkmark &\checkmark & &  ~ & 35.5 \textcolor{tablegreen}{\small (\textbf{+1.3)}} & 56.4 & 35.3\\
~ &\checkmark & &\checkmark & ~ & 35.1 \textcolor{tablegreen}{\small (\textbf{+0.9)}} & 55.5 & 35.6\\
~ &\checkmark &\checkmark &\checkmark &~ & 36.1 \textcolor{tablegreen}{\small (\textbf{+1.9)}} & 57.6 & 35.6\\
\bottomrule
\end{tabular}
} 
\caption{The effect of the proposed pipeline, DFP, and TAL. 'Off AP' means the corresponding AP using the base detector on the offline setting. 'Pipe.' denotes the proposed pipeline, marked in \colorbox{baselinecolor}{gray}, while '$\uparrow$' indicates the corresponding improvements. '\textcolor{tablegreen}{($\cdot$)}' indicates the relative improvements based on the strong pipeline.} 
\label{tab:table3}
\end{table}

\begin{table}[t]
\vspace{-.2em}
\centering
\subfloat[
$\nu > 1$.
\label{tab:table41}
]{
\scalebox{0.72}{
\begin{tabular}{@{}c|c|c|ccccc@{}}
\toprule
$\tau$ & $\nu$ & sAP & $\rm sAP_{50}$ & $\rm sAP_{75}$ & $\rm sAP_{s}$ & $\rm sAP_{m}$ & $\rm sAP_{l}$\\
\midrule
0.2 & 1.3 & 35.6 & 57.1 & 36.0 & 13.0 & 36.9 & 62.8\\
0.2 & 1.4 & 35.8 & 57.4 & 35.0 & 13.2 & 37.0 & 62.5 \\
0.2 & 1.5 & 35.8 & 57.2 & 35.4 & 12.8 & 36.8 & \textbf{63.6}\\
0.3 & 1.3 & 35.8 & 56.8 & 35.3 & 13.7 & 36.3 & 62.6\\
0.3 & 1.4 & \textbf{36.1} & \textbf{57.6} & 35.6 & 13.8 & \textbf{37.1} & \textbf{63.3}\\
0.3 & 1.5 & 35.9 & 57.6 & 35.0 & 13.2 & 36.8 & 63.3\\
0.4 & 1.3  & 35.3 & 56.9 & 35.2 & 12.8 & 35.1 & 61.6\\
0.4 & 1.4 & 35.7 & 57.1 & \textbf{36.0} &  13.2 & 36.6 & 61.0\\
0.4 & 1.5 & 35.4 & 56.7 & 35.3 & 13.6 & 35.9 & 61.8\\
\bottomrule
\end{tabular}} 
}
\centering
\hspace{0.5em}
\subfloat[
$\nu < 1$.
\label{tab:table42}
]{
\scalebox{0.72}{
\begin{tabular}{@{}cc@{}}
\toprule
$\nu$ & sAP\\
\midrule
1.5 & 35.9\\
1.4 & \textbf{36.1}\\
1.3 & 35.8\\
1.2 & 35.8\\
1.1 & 35.5\\
1.0 & 35.6\\
0.9 & 35.4\\
0.8 & 35.4\\
0.7 & 35.0\\
\bottomrule
\end{tabular}
} 
}
\\
\centering
\caption{Grid search of $\tau$ and $\nu$ in Eq.~\ref{eq:eq2} for TAL.}
\label{tab:table4}
\end{table}

\paragraph{Value of $\tau$ and $\nu$.} As depicted in Eq.~\ref{eq:eq2}, the value of $\tau$ acts as a threshold to monitor newly emerging objects while $\nu$ controls the degree of attention on the new objects. We set $\nu$ larger than 1.0 so that the model pays less attention to the new-coming objects. We conduct a grid search for the two hyperparameters in Tab.~\ref{tab:table41}, where $\tau$ and $\nu$ achieve the best performance at 0.3 and 1.4 respectively. When $\nu$ is less than 1, we will pay more attention to new-coming objects and decrease the performance as shown in Tab.~\ref{tab:table42}.



\subsection{Further Analysis}
\paragraph{Robustness at different speeds.}
We further test the robustness of our model at different moving speeds of the driving vehicle. To simulate the static (0$\times$ speed) and faster speed (2$\times$) environments, we re-sample the video frames to build new datasets. For 0$\times$ speed setting, we treat it as a special driving state and re-sample the frames to the triplet $(F_{t}, F_{t}, G_{t})$. It means the previous and current frames have no change and the model should predict the non-changed results.
For 2$\times$ speed setting, we re-build the triplet data as  $(F_{t-2}, F_{t}, G_{t+2})$. This indicates the faster moving speed of both the ground truth objects and the driving vehicle. 

Results are listed in Tab.~\ref{tab:table5}. For 0$\times$ speed, the predicting results are supposed to be the same as the offline setting. However, if we only adopt the basic pipeline, we can see a significant performance drop (-1.9 AP) compared to the offline, which means the model fails to deduce the static state. By adopting the DFP module into the basic pipeline, we recover this reduction and achieve the same results as the offline performance. It reveals that DFP, especially the static flow, is a key to extracting the right moving trend and assisting in prediction. It is also worth noting that at 0$\times$ speed, all the weights in TAL are one thus it has no influence. For 2$\times$ speed, as the objects move faster, the gap between offline and streaming perception is further expanded. Meanwhile, the improvements from our models, including the basic pipeline, DFP, and TAL, are also enlarged. These robustness results further manifest the superiority of our method.  

\begin{table}[t]
\centering
\scalebox{0.75}{
\begin{tabular}{@{}c|ccc|c|ccc@{}}
\toprule
Model & Pipe. & DFP & TAL & Off AP & $\rm sAP_{0x}$ & $\rm sAP_{1x}$ &  $\rm sAP_{2x}$\\
\midrule
\multirow{4}*{YOLOX-L} &   &  &  & \multirow{4}*{38.3} & 38.3 & 31.2 & 24.9 \\
~ &  \checkmark &  &  & ~ &36.4 & 34.2 & 31.3 \\
~ &  \checkmark & \checkmark &  & ~ &38.3 & 35.5 & 32.9 \\
~ &  \checkmark & \checkmark & \checkmark & ~ & 38.3 & 36.1 & 33.3 \\
\bottomrule
\end{tabular}
} 
\caption{Results on different moving speed settings. The 0$\times$ static setting actually equals to the offline setting. Subscripts indicate different moving speeds.} 
\label{tab:table5}
\end{table}

\begin{table}[t]
\centering
\scalebox{0.8}{
\begin{tabular}{@{}c|cc|c@{}}
\toprule
Forecasting manner &  $\rm sAP_{1x}$ & $\rm sAP_{2x}$ & Extra Latency\\
\midrule
 Offline Det   &  31.2 & 24.9 & 0 ms\\
KF Forecasting &  35.5 & 31.8 & 3.11 ms\\
 Ours (E2E) &  \textbf{36.1} & \textbf{33.3} & \textbf{0.8 ms}\\
\bottomrule
\end{tabular}
} 
\caption{Comparison results on different forecasting manners.} 
\label{tab:rebuttal_prediction}
\end{table}

\begin{table}[t]
\centering
\resizebox{\linewidth}{!}{ 
\begin{tabular}{@{}c|ccc|ccc@{}}
\toprule
Method & sAP & $\rm sAP_{50}$ & $\rm sAP_{75}$ & $\rm sAP_{s}$ & $\rm sAP_{m}$ & $\rm sAP_{l}$\\
\midrule
\multicolumn{7}{c}{\textbf{Non-real-time methods}}\\
\midrule
Streamer (size=900)\cite{streamer} & 18.2 & 35.3 & 16.8 & 4.7 & 14.4 & 34.6\\
Streamer (size=600)\cite{streamer} & 20.4 & 35.6 & 20.8 & 3.6 & 18.0 & 47.2\\
Streamer + AdaScale\cite{adascale,adaptivestreamer} & 13.8 & 23.4 & 14.2 & 0.2 & 9.0 & 39.9\\
Adaptive Streamer\cite{adaptivestreamer} & 21.3 & 37.3 & 21.1 & 4.4 & 18.7 & 47.1\\
\midrule
\multicolumn{7}{c}{\textbf{Real-time methods}}\\
\midrule
1st place (size=1440)$^\dagger$\cite{yoloxx} & 40.2 & 68.9 & 39.4 & 21.5 & 42.9 & 53.9\\
2nd place (size=1200)$^\dagger$\cite{2nd} & 33.2 & 58.6 & 30.9 & 13.3 & 31.9 & 40.0\\
Ours-S & 28.8 & 50.3 & 27.6 & 9.7 & 30.7 & 53.1\\
Ours-M & 32.9 & 54.0 & 32.5 & 12.4 & 34.8 & 58.1\\
Ours-L & 36.1 & 57.6 & 35.6 & 13.8 & 37.1 & 63.3\\
Ours-L (size=1200)$^\dagger$ & 42.3 & 64.5 & 46.4 & 23.9 & 45.7 & 68.1\\
\bottomrule
\end{tabular}
} 

\caption{Performance comparison with state-of-the-art approaches on Argoverse-HD dataset. Size means the shortest side of input image and the input image resolution is 600$\times$960 for our models. '$\dagger$' means using extra dataset and TensorRT.} 
\label{tab:table6}
\end{table}

\begin{figure*}[t]
\setlength{\abovecaptionskip}{0pt}
\begin{center}
\includegraphics[width=0.9\linewidth]{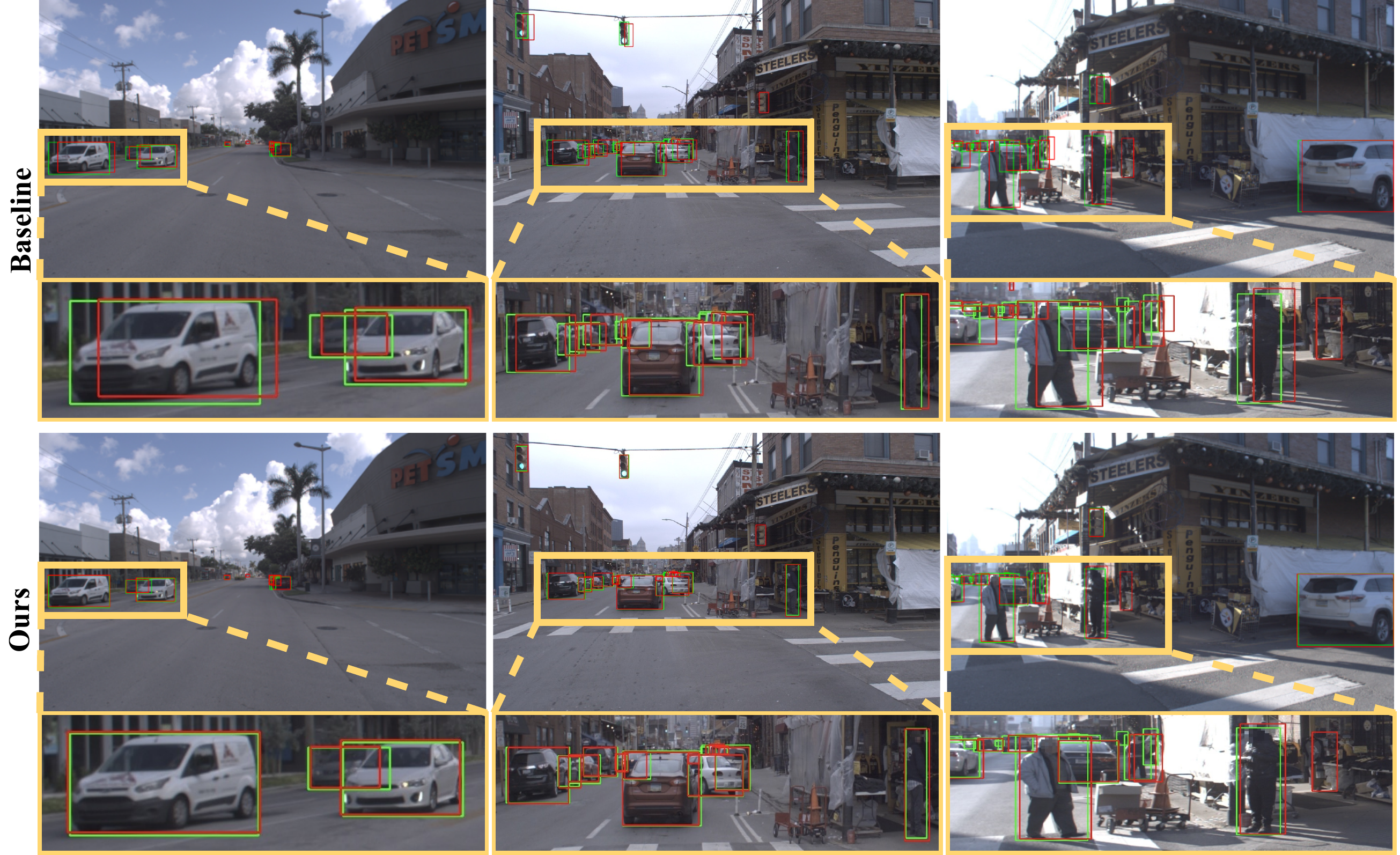}
\end{center}
\caption{Visualization results of the baseline detector and the proposed method. The green boxes represent ground truth boxes, while red ones represent prediction results.}
\vspace{-0.4cm}
\label{fig:visualization}
\end{figure*}

\paragraph{Comparison with Kalman Filter based forecasting.} We follow the implementation of \cite{streamer} and report the advanced baseline of Kalman Filter based forecasting in Tab.~\ref{tab:rebuttal_prediction}. For ordinary sAP (1$\times$), our end-to-end method still outperforms the advanced baseline by 0.5 AP. Further, when we simulate and evaluate them with faster moving (2$\times$), our model shows more superiority of robustness (33.3 sAP v.s. 31.8 sAP). Besides, our model brings less extra latency (0.8 ms v.s. 3.1 ms taking the average of 5 tests).

\paragraph{Visualization results}
As shown in Fig.~\ref{fig:visualization}, we present the visualization results. For the baseline detector, the predicting bounding boxes encounter severe time lag. The faster the vehicles and pedestrians move, the larger the predictions shift. For small objects like traffic lights, the overlap between predictions and ground truth becomes small and is even non. In contrast, our method alleviates the mismatch and fits accurately between the predicting boxes and moving objects. It further confirms the effectiveness of our method.

\paragraph{Comparison with state-of-the-art.}
We compare our method with other state-of-the-art detectors on Argoverse-HD dataset. As shown in Fig.~\ref{tab:table6}, real-time methods show absolute advantages over non-real-time detectors.
We also report the results of the 1st and 2nd place in Streaming Perception Challenge. They involve extra datasets and accelerating tricks, while our methods get competitive performance and even surpass the accuracy of the 2nd place without any tricks. Once we adopt the same tricks, our method outperforms the 1st place by a significant margin (2.1 sAP).

\section{Conclusion}
This paper focuses on a streaming perception task that takes the processing latency into account. Under this metric, we reveal the superiority of using a real-time detector with the ability of future prediction for online perception. We further build a real-time detector with Dual-Flow Perception module and Trend-Aware Loss, alleviating the time lag problem in streaming perception. Extensive experiments show that our simple framework achieves state-of-the-art performance. It also obtains robust results on different speed settings. We hope that our simple and effective design will motivate future efforts in this practical and challenging perception task.

\paragraph{Limitations}
In real-world scenarios, the assumption of real-time processing may be violated due to limited hardware resources or high-resolution input. Moreover, we can see that the gap between \emph{offline} setting and our model of online perception still exists by a large margin, indicating that there is still unexplored room for streaming perception.

\paragraph{Acknowledge} This paper is supported by the National Key R\&D Plan of the Ministry of Science and Technology (Project No. 2020AAA0104400). It was also funded by China Postdoctoral Science Foundation (2021M690375) and Beijing Postdoctoral Research Foundation.       

{\small
\bibliographystyle{ieee_fullname}
\bibliography{streamyolo}
}

\end{document}